\documentclass[letterpaper]{article} 
\usepackage{aaai25}  
\usepackage{times}  
\usepackage{helvet}  
\usepackage{courier}  
\usepackage[hyphens]{url}  
\usepackage{graphicx} 
\usepackage{natbib}  
\usepackage{caption} 
\usepackage{algorithm}
\usepackage{algorithmic}
\usepackage{booktabs}

\usepackage{newfloat}
\usepackage{listings}
\DeclareCaptionStyle{ruled}{labelfont=normalfont,labelsep=colon,strut=off} 
\floatstyle{ruled}
\newfloat{listing}{tb}{lst}{}
\floatname{listing}{Listing}
\title{Operator Feature Neural Network for Symbolic Regression}
\author{
    Yusong Deng,
    Min Wu,
    Lina Yu,
    Jingyi Liu,
    Shu Wei,
    Yanjie Li,
    Weijun Li
}
\affiliations{
    \textsuperscript{\rm 1}Institute of Semiconductors, Chinese Academy of Sciences\\

    dengyusong22@mails.ucas.ac.cn
}

\usepackage{bibentry}

\begin{document}

\maketitle

\begin{abstract}
Symbolic regression is a task aimed at identifying patterns in data and representing them through mathematical expressions, generally involving skeleton prediction and constant optimization. Many methods have achieved some success, however they treat variables and symbols merely as characters of natural language without considering their mathematical essence. This paper introduces the operator feature neural network (OF-Net) which employs operator representation for expressions and proposes an implicit feature encoding method for the intrinsic mathematical operational logic of operators. By substituting operator features for numeric loss, we can predict the combination of operators of target expressions. We evaluate the model on public datasets, and the results demonstrate that the model achieves superior recovery rates and high $R^2$ scores. With the discussion of the results, we analyze the merit and demerit of OF-Net and propose optimizing schemes.
\end{abstract}

\section{Introduction}
With the rapid advancement of deep learning, artificial intelligence algorithms have achieved satisfactory outcomes in various tasks within computer vision and natural language processing. A faction of scientists has endeavored to utilize artificial intelligence methodologies for knowledge discovery, specifically referring to the extraction of symbolic expressions from numerical data to map its patterns, a category of tasks known as symbolic regression.

Up to the present, foundational and mainstream methods of symbolic regression can primarily be categorized into three directions. Symbolic Physics Learner (SPL) \cite{sun2022symbolic} utilizes the methods of Monte Carlo tree search to discover expressions. Another approach, known as the Deep Symbolic Regression (DSR) \cite{petersen2019deep}, employs the strategy gradient of reinforcement learning to guide neural networks searching in the solution space. These methods has served as the baseline for many subsequent improvements \cite{mundhenk2021symbolic, li2024discovering} based on solution space searching. Another notable direction, represented by Neural Symbolic Regression that Scales (NeSymReS) \cite{biggio2021neural}, centralizes on leveraging the learning and fitting capabilities of neural networks to generate predicted candidates end-to-end. Specific enhancements to this category of methods have been made by many researches \cite{kamienny2022end, vastl2024symformer, meidani2023snip, wu2023discovering, li2022transformer}, achieving commendable results. Furthermore, Silviu-Marian Udrescu \cite{udrescu2020ai1, udrescu2020ai2} utilizing  neural networks' fitting capabilities as the approximation of target expressions, leveraging the automatic differentiation mechanism to discover the structure between varying trends of different variables, which has been used to infer properties such as translational symmetry and additive separability, thereby facilitating the dissection and piecewise parsing through variable separation. Additionally, numerous other methods \cite{landajuela2022unified, xu2023rsrm, holt2023deep, liu2023snr, chen2023transformer}, integrating the strengths of baseline methods, make more detailed enhancements.

However, in the vast majority of these approaches, different operator symbols, such as '+' and '×', are merely treated as two characters of equal operational status during the prediction. Though the divergence in symbols can, to a certain extent, influence algorithm parameters via backward propagation, thus reflecting the differences between symbols, such disparities can intermingle and lack precision. For instance, in the case of $y = e^{x_1 + x_1^2 + x_2^3}$, assuming the predicted outcome is $y = e^{x_1 + x_1^2 \times x_2^3}$, even if the content and sequence of each characters in predicted expression are entirely accurate except only '+' character predicted incorrectly as '$\times$', this could lead to a significant numerical loss, which is then distributed dispersedly among all characters. We can evidently see that equating symbolic expressions directly to natural language characters is irrational.

Silviu-Marian Udrescu's methods \cite{udrescu2020ai1, udrescu2020ai2} proposed to simplify expressions by determining the properties among variables, which indeed notably focuses on the fundamental mathematical logic inherent to the differing operators like '+' or '×'.  However, these methods necessitates that the variables can be segregated, specifically that they remain mutually independent within each segment. Their stringent explicit judgments have confined its application, rendering it ineffective when confronted with highly intertwined expressions. Hence, invigorated by the quintessential principles of these methods, we employ the implicit feature representation of symbolic operators, utilizing soft constraints and loosely-coupled relationships, in association with deep operator networks (DeepONets) \cite{lu2019deeponet}, feature encoders, and other techniques, endeavoring to find the underlying functional operational logic inherent to unknown expressions from different known operators. Pertinent details of related neural networks will be expounded upon in 'Related Work', whereas a more precise algorithmic process will be elucidated in 'Method'.

\section{Related Work}

\subsection{DeepONets}

It is presented by T. Chen \cite{chen1995universal} that a neural network with single hidden layer can theoretically approximate arbitrary nonlinear continuous operator, which refers to a mapping from one function space to another. Building upon this foundation, considering neural network optimization on generalization errors, Lu leads to the proposition of Deep Operator Networks (DeepONets) \cite{lu2019deeponet}, which consists of a trunk net and a branch net, structured in a manner consonant with the format of the approximation theorem. 

The branch net accepts the original function's features as input and outputs new features of target function transformed by the operator. It is essentially a mapping of different feature fields, representing the process of the operator acting on the function. Simultaneously, the trunk net takes the coordinates in the space of target function as inputs and outputs the encoded positional features. The function feature from branch net and the position feature from the trunk net together compute the value of target function at the respective coordinates. This process that culminating in an effective approximation of target function is regarded as the fitting procedure for the operator.

Numerous researchers have subsequently made further modifications \cite{lu2022comprehensive, lu2022multifidelity, jin2022mionet, mao2021deepm} based on this foundation, which yet are mainly tailored towards specific issues in designated scenarios like equations coupling or multiple function space, offering considerable reference value when addressing related problems with the core concept and fundamental framework remaining unchanged. Operator networks were initially designed for dynamic systems and problems related to partial differential equations and have been following this trajectory of development. Nonetheless, it is evident that they can be readily applied to simpler symbolic operators as encountered in elementary mathematics. 

\subsection{Discrimination and Multimodal}

The process of transitioning from symbolic expressions to operator architectures can largely be understood as a multi-target discrimination task. Specifically, this involves transforming the issue of determining which operators appear in a particular place into a discrimination problem of identifying the presence of the interaction of each operator, thus necessitating the neural network to perform judgment actions. Since the introduction of the transformer \cite{vaswani2017attention} architecture, centered around the attention mechanism, it has garnered extensive and widespread recognition. The encoder part of the transformer was isolated as the bidirectional encoder representations from transformers (BERT) \cite{devlin2018bert} architecture was proposed for feature extraction and processing which has been extensively utilized in tasks such as text classification as well as other long sequence discrimination challenges. Due to its effective performance and broad applicability, this structure has been adopted in a variety of fields including vision for classification, regression, and other tasks, achieving notable results.

Moreover, when handling multimodal data, the feature encoders and decoders effectively accomplish the task of mapping features between different domains. In the context of symbolic regression, the set-transformer \cite{lee2019set} is widely employed as an encoder. It is capable of accepting input-output pairs, reducing the computational scale, and ensuring that the order of data points does not affect the outcome. 

\section{Methods}

\subsection{Directed graph for symbolic expression representation}

Presently, the majority of symbolic regression methods employ tree representation, in which each node denotes a character and the edges symbolize the sequential relationships among characters, typically arranged in preorder traversal. This approach is facing with two issues.

One is that for a given expression, selecting various nodes as the root or different sequence may lead to different structures but representing for a semantically equivalent tree. This is profoundly influential for neural networks. Given a network without random, under the circumstances of determined parameters and unchanged inputs, a specific output will certainly be generated. If training input associated with multiple output labels, the neural network's parameter optimization during training will target multiple objectives. Each direction will be influenced by other objects, causing deviation. The overall loss would be a trade off among various labels, hindering the accurate capture of the correct results.

Another predicament is that the prediction of trees utilizes the logic of natural language represented by strings. Despite the string, that forms the tree through preorder traversal, partially reflects some character features to a certain extent, this feature is imprecise in mathematics as mentioned earlier. Therefore, instead of interpreting the expressions through the character sequence in natural language, we place our focus on the mathematical operating logic beneath the expression. For this purpose, we have revamped the way of expression representation.

\begin{figure}[h]
\centering
\includegraphics[width=0.9\columnwidth]{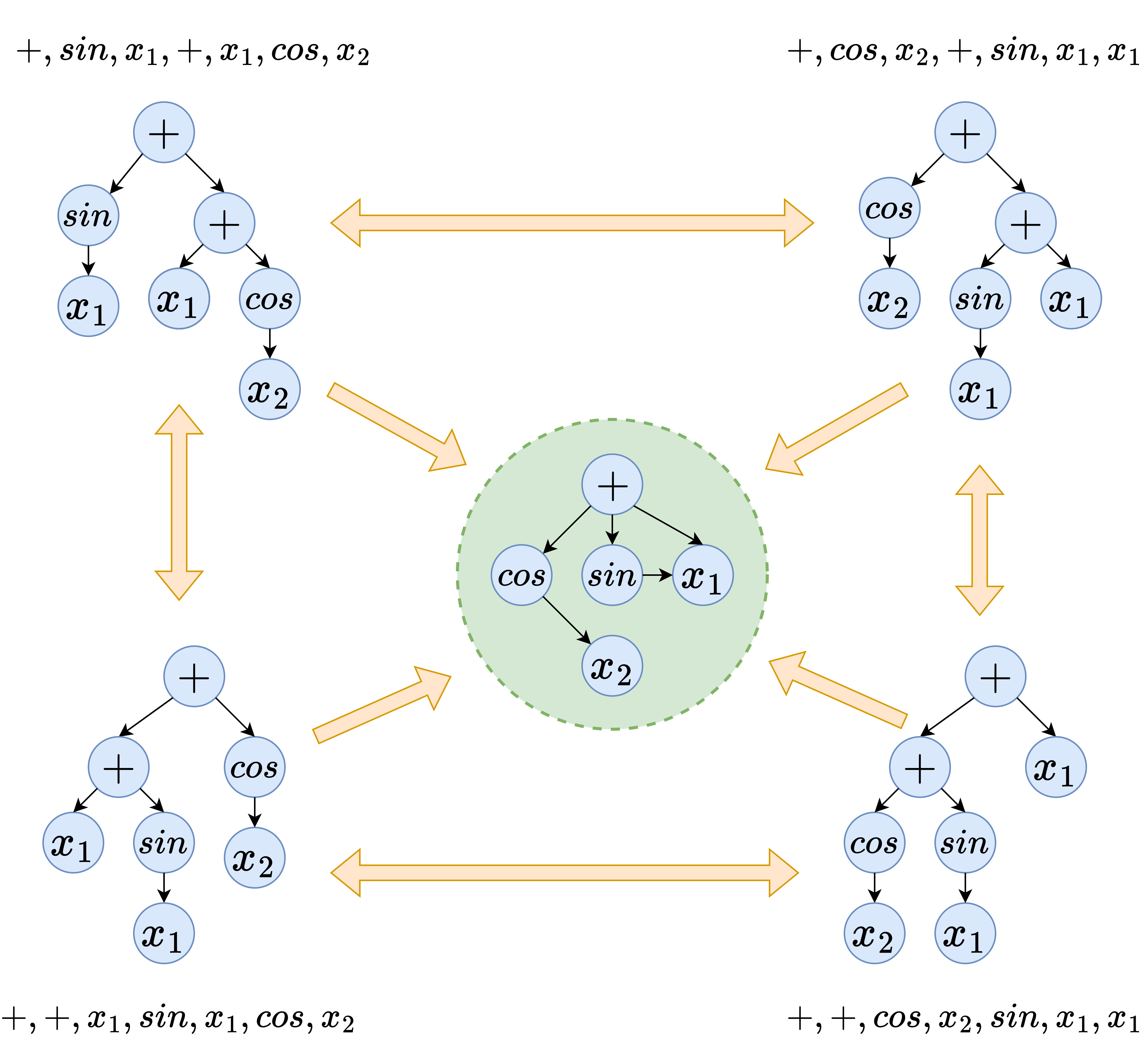}
\caption{A example for tree and graph. The expression $y = sin(x_1) + cos(x_2) + x_1$ has multiple preorder encoding for tree but unique for graph.}
\label{Expressions}
\end{figure}

We utilizing directed graphs as our foundational framework where the nodes are categorized into two distinct types: variable nodes and operator nodes. Variable nodes denote the functions being inputted, and are defined as $ f = x_i $ explicitly in symbolic regression problems. To streamline the overall operator structure, we adopt $ f = c \times x_i $ as our elemental input function, where 'c' symbolizes a constant to be optimized, thereby broadening the expressive range of the operator under equivalent scale constraints. Operator nodes represent the operators that act upon functions, with the transformed functions as outputs. Expressions are realized through the sequential combination of multiple operators with connections between operators denoted by the edges.

We employ the adjacency matrix form to denote the one-hot encoding of operator graphs. label $[x,y]=1$ implies the existence of operator x acting on the output of operator y, where certain pieces of data flow from operator y to operator x. At the same time, this representation method resolves the issue of multiple encodings corresponding to a single expression. Although the process of decoding expressions from their corresponding encodings may introduce redundant skeleton candidates, due to the fact that the process of mapping concrete expressions from adjacency matrices isn't bijection, this can be partly resolved via search algorithms. Hence, this graphical representation still solves the issues encountered with tree representation to a certain extent, greatly reducing the amount of redundant and useless possibilities.

\begin{table}[h]
\centering
\begin{tabular}{lcc}
\toprule
Number & Operator & range\\
\midrule
1 & $G(u_1,u_2)=c\times u_1 + u_2$ & $(-\infty,+\infty)$ \\
2 & $G(u_1,u_2)=c\times u_1 \times u_2$ & $(-\infty,+\infty)$ \\
3 & $G(u)=u^{-1}$ & $(-\infty,t) and (t,+\infty)$ \\
4 & $G(u)=sin(u)$ & $(-\infty,+\infty)$ \\
5 & $G(u)=cos(u)$ & $(-\infty,+\infty)$ \\
6 & $G(u)=exp(u)$ & $(-t,t)$ \\
7 & $G(u)=pow(u,c),c>1$ & $(-t,t)$ \\
8 & $G(u)=pow(u,c),0<c<1$ & $[0,+\infty)$ \\
9 & $G(u)=log(u)$ & $(t,+\infty)$ \\
10 & $G(u)=u + c$ & $(-\infty,+\infty)$ \\
11 & $G(u)=u\times c$ & $(-\infty,+\infty)$ \\
\hline
\end{tabular}
\caption{The operator set where '$t$' represents the threshold and '$u$' refers to $u(x_n)$, a function of variable $x_n$. The range is for input $u$ so that the output $G(u)$ is limited.}
\label{Operator Set}
\end{table}

In an ideal scenario, a basic operator set is complete and independent that can perform any function representable in mathematics through nested combinations with each mathematical expression corresponds to only one operator combination structure. In our established operator set, for instance, operator 2 for multiplication and operator 7 for power ($c>1$) are dependent as the square of X is the same as x times x. Yet, these two operators cannot be substituted in most instances, like $sin(x) \times sin(1.5x)$, which will miss a part of the potential solution space if they are substituted. Only when both sides are completely consistent, including constant optimization, can operator 2 be represented by operator 7. But this is obviously impractical, because constant optimization exists after the expression skeleton. The reason why we choose not to use multiplication and pow ($0<c<1$) to replace the pow ($c>1$) is because for slightly complex functions, this replacement will introduce a large number of candidates and constants, leading to an unacceptable burden for path searching and constant optimization. Therefore, we make a trade-off between operator independence, completeness and search complexity, forming our operator set in table 1, which can represent most mathematical formulas without extensive repetitious presentation while decreasing search difficulty. Increase of thresholds is effective in solving the overlap of operators.Naturally, this is likely not the optimal choice, but it is basically sufficient.

\subsection{The Network}

\begin{figure*}[t]
\centering
\includegraphics[width=0.9\textwidth]{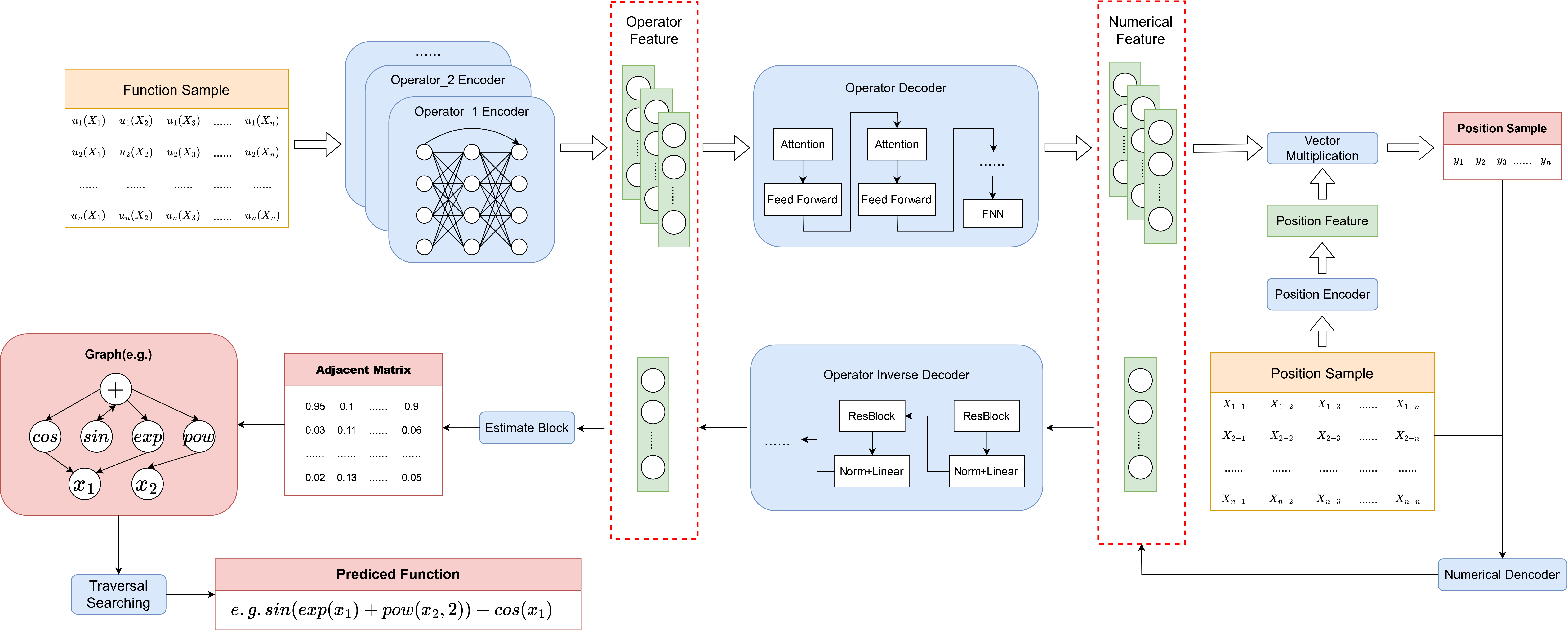}
\caption{The structure of OF-Net. The forward process of operator fitting is shown by hollow arrows and the linear arrows represents for the backward inference.}
\label{OF-Net}
\end{figure*}

The network's structure is shown in figure 2, which can be divided into forward fitting and backward inference. 

The forward network employs the basic infrastructure of the operator network, accomplishing an overall fit of each operator's computational process and capturing operator features intermediately in the branch net. The data sampled from positions are projected into a positional feature through a position encoder, corresponding to the trunk net. The branch network component includes distinct operator encoders for each operator and just one shared operator decoder for every different operators. The function samples pass sequentially through the operator encoder and decoder, and the resulting vector multiply with position features getting the final numerical fitting outcome. This process constitutes the network's fitting for a single operator, and upon successful fitting, the network is considered capable of representing the target operator.

The trunk net is responsible for positional encoding is solely dependent on the input positions and is unaffected by the input function or the target operator. Consequently, the positional encoding remains consistent across different operators. However, the target operator's variation leads to differing numerical results at the same coordinates, necessitating that only the branch network represents the unique characteristics of each operator. The use of both an encoder and a decoder in the branch network is due to the fact that the final output of the branch net possesses the ability to fit the operator under the given positional encoding, representing the numerical fitting characteristics, which we refer to as numerical features. What we require, however, is the structural characteristic in terms of mathematical operations, known as operator features. Therefore, a mapping in the feature space is necessary. Employing a single, shared decoder ensures that for different operators, only the operator encoder varies. This approach concentrates the differences between operators within the encoder, allowing the encoder to extract operator features that encapsulate the unique operational logic, thereby more effectively highlighting their distinct attributes.

The backward network is responsible for deriving the final predictive expression from the basic and target operator features. Notably, for an unknown expression, we only have input-output pairs and a value range. Therefore, the ideal process involves fixing the operator decoder and position encoder while using known data to train an operator encoder for the target operator, allowing it to fit the final numerical labels, before the next step of prediction. However, this process is time-consuming that impractical to retrain a network module for each new operator. To resolve this, we add two modules to facilitate the reverse inference process of the forward network. The numerical decoder retrieves numerical features given known positional features and numerical labels, while the operator inverse decoder maps from the numerical feature space to the operator feature space. This allows us to use a forward propagation of the backward network instead of reverse training to obtain operator features. The judgment network takes the basic and target operator features as input and predicts their relationship, outputting an adjacency matrix. The search module, not actually part of the neural network, is a search strategy to constructs functions by searching paths in operator adjacency matrix. We balance search completeness and efficiency so that most common functions are discovered earlier. Details will be introduced in 'Searching'.

The judgment block is based on BERT due to its intrinsic attention mechanism that capably discerns the relation of diverse operator, with all features treated equitably and without unidirectional perception masks. Another module impervious to the sequence of inputs is the numerical decoder, where the order of sample points should not bias the function's interpretation, therefore we employ a set-transformer with canonical structure. The operator decoder also utilizes a transformer framework, which eschews the typical practice of feeding different sequence elements as input and instead maps one operator feature to multiple spaces, aggregating the mapped features to enrich the operator feature with multidimensional information. The judgment block and numerical decoder employ standard stacked modules while remaining sections include deep fully connected networks with residual links, arranged alternately with single layers and residual blocks. Both the network scale and feature lengths are adjustable hyper-parameters.

\subsection{Searching}

The search referenced here diverges from algorithms like DSR and does not aim to navigate a vast solution space directly toward a target. Instead, guided by an adjacency matrix, it adopts a traversal method like breadth-first search within a constrained area,.

The search process is directed by the pre-obtained operator adjacency matrix, significantly narrowing the solution space. Consequently, the approach shifts away from random exploration towards a pseudo-traversal method designed to preserve more completeness. This "pseudo" derives from the strategy not being a comprehensive exploratory search like depth-first search but one that discards highly uncommon solution spaces to more quickly access and identify frequent structures.

Firstly, we impose restrictions on the nesting of operators: the duplication of trigonometric, power, exponential, and logarithmic operators within their sub-nodes is restricted, which is a common constraint in symbolic regression to maintain the simplicity of expressions, as most conventional expressions do not involve nesting of these operators or possess uncomplicated equivalent forms without nesting. For addition and multiplication operators, however, there is an upper limit to the number of times their sub-nodes may duplicate themselves. This measure is adopted because certain common expressions inherently contain loops, exemplified by $y = cos(x_1) + cos(x_1 + x_2)$, the potential infinite loop between addition operators and cosine operators without nesting limits.

We set $c \times x$ as the unique initial variable, designating it as a leaf in the computational graph. Once this node is encountered during the search, that branch of the search is terminated and no further extension is performed. Other operators, serving as non-leaf nodes, will sequentially act on the initial variable. When searching for candidate expressions from the graph, the root must first be determined. We evaluate the potential of each operator as the root based on its out-degree and in-degree. An operator with an in-degree of zero and a non-zero out-degree is most likely to be the root and will be prioritized. If an operator has a non-zero in-degree but an out-degree greater than its in-degree, it may also serve as a root, and this category of operators will be considered when the search has not yielded an ideal solution. The greater the difference between the out-degree and in-degree, the more likely the operator is to become the root of the target expression, which is merely based on empirical observations rather than rigorous mathematical measure of probability.

Meanwhile, we have made adjustments to the situation that adjacency matrix has errors, ensuring that the search returns relatively accurate candidate expressions without collapse. When encountering operators at non-leaf nodes without out-degree, the strategy dictates returning an empty set or the initial variable. In instances where multiple operators lack in-degree, we sequentially initiate the search with these operators until completion or until an expression is identified. With addition or multiplication operators receiving multiple inputs, it can be unclear whether they truly represent the addition of multiple terms or if some inputs are erroneously interpreted.  Consequently, we opt to first search for the addition of two terms, incorporating a placeholder for potential expansion. This placeholder is disregarded during subsequent optimization of constants, yet should no suitable expression emerge in a given round, we extend one term at the placeholder's position, conduct an independent search, and integrate the outcome into the original candidate expression. To prevent indefinite expansion of placeholders, we set a maximum number of e expansions as a hyperparameter.

\begin{figure*}[t]
\centering
\includegraphics[width=0.9\textwidth]{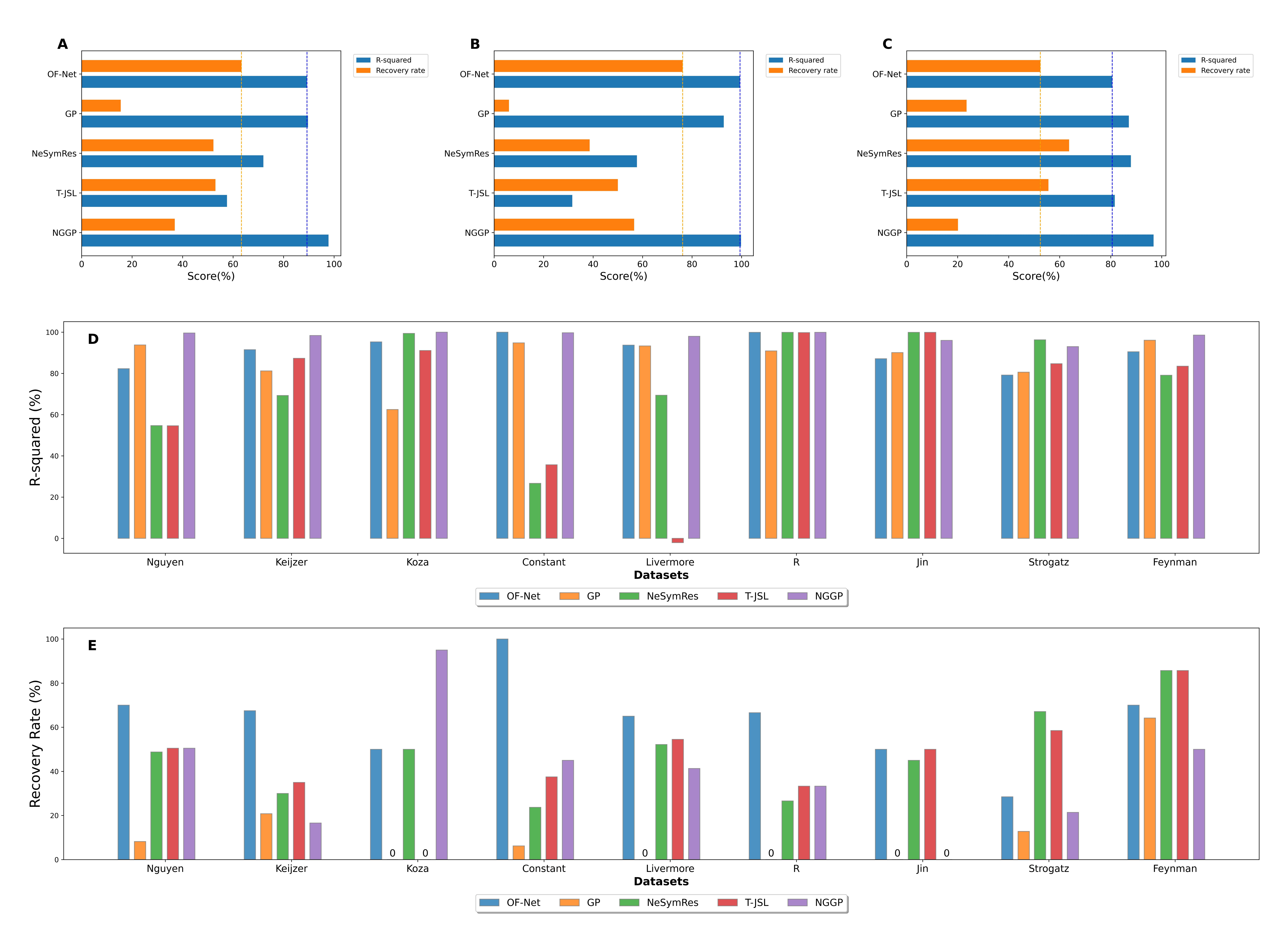}
\caption{Statistical result of the experiment. A: performance on holistic data. B: performance on univariate data. C: performance on bivariate data. D: $R^2$ on different datasets. E: recovery rate on different datasets}
\label{Result}
\end{figure*}

Constant optimization is not conducted all the time, but occurs after generating a list of candidate skeletons in each round. Skeletons are added sequentially to the list according to the order so that it suffices to simply traverse the list in sequence for constant optimization. We employ the classical BFGS method for the constants optimization where 80 points are randomly selected as reference points. Each candidate expression undergoes constant initialization 30 times, with the first ten in the range (0, 1) and the subsequent twenty in the range (-10, 10), while the range for operator $pow(f, c) (0<c<1)$ remains consistently within (0, 1). It's noteworthy that for the pow operator, the BFGS algorithm struggles to effectively optimize constants, as constants are represented in floating-point format. When the exponent includes a constant and the base might be negative, complex numbers are introduced, undermining the relevance of the $R^2$ because an incorrect complex value's $R^2$ can significantly exceed 1. Even though we recognize that legitimate expressions do not encompass such cases, it's impossible to guarantee the absence of complex numbers during random initialization and the computational process.

To address this challenge, we explored two alternative strategies. If it's assumed a priori that the expression does not contain floating-point exponent terms, one method involves iteratively trying different integer. This approach reliably captures the correct skeleton but at the cost of introducing an extensive amount of futile computations for expressions with higher-order terms, thus wasting considerable time. Another method involves respectively modulating the subtraction in Mean Squared Error (MSE) and $R^2$ during numerical optimization and assessment, substituting square with euclidean norm. This can enhance the precision of constant optimization to a certain degree. However, empirical tests showed that the improvement in precision is limited, and some expressions still fail in constant optimization even though their skeleton is accurate. Therefore, we apply different optimization strategies tailored to various expressions, with detailed discussions in  'Experiments and Discussion'.

\section{Experiments and Discussion}

\subsection{Conditions and Setup}

The operator feature length is set at 500, numerical feature length at 200, with 200 function sampling points in branch net and 1600  position sampling points in trunk net. The position encoder consists of univariate block and bivariate block that each incorporates six single hidden layers and five residual blocks, with a maximum layer length of 800. Both the operator encoder and the operator inverse decoder are structured with five single hidden layers and four residual blocks, with a maximum layer length of 1000. The operator decoder and judgment block each utilize a stack of three attention layers. We utilize complex constant optimization and integer traversal optimization respectively when the exponent is greater and is not greater than 5 for univariate expressions, while 3 serves as the threshold for bivariate expressions.

We compared our method against four different approaches, denoted as Genetic Programming (GP) for searching, NeSymReS and T-JSL\cite{li2022transformer} for pre-trained, and DSO/NGGP\cite{mundhenk2021symbolic}, which is known as SOTA, for combined model. We set pop size at 1000 and generation at 40 in GP and utilize standard configuration of others. The test set includes univariate and bivariate expressions in SRBench and the majority of the known public datasets currently available. We believe that recovery rate is one of the most significant indicators of performance in symbolic regression, as achieving complete recovery and discovering the true formula aligns with the fundamental goals of symbolic regression. However, due to the involvement of constant fitting, we approached the issue from a numerical perspective, making the $R^2$ score indispensable. An $R^2$ score of 0.999 and 0 does not imply equivalent algorithm performance. Therefore, we tested each expression 10 times, recording the $R^2$ scores and recovery rates. If complete recovery was achieved at least once, we recorded the $R^2$ as 1; otherwise, we averaged the two median values. All experiments were conducted using the same hardware configuration: NVIDIA Corporation GV100 [TITAN V].

\subsection{Result}

The experimental results are illustrated in figure 3, with detailed outcomes in appendices. Notably, due to the introduction of random in most methodologies, such as the initialization of BFGS constant optimization and position sampling, as well as potential variations in the runtime environment, some discrepancies between our test results and the descriptions in original articles are reasonable. The results we present were retested under uniform software and hardware conditions.

In evaluating whether a prediction is completely recovered, we first consider the $R^2$ value. If $R^2$ exceeds 0.999999, we believe that this prediction has the possibility to be correct. Subsequently, for predictions meeting the $R^2$ criterion, we conduct a manual verification, where we manually calculate and compare the predicted expression with the correct label to determine a complete recovery. Therefore, even if the predicted skeleton is accurate but the constant optimization is inadequate, we do not consider it to be completely recovered.

Experiment result indicates that OF-Net achieved the highest recovery rate across the holistic dataset, an outcome anticipated given its design tailored specifically for operator features. OF-Net can almost invariably reconstruct the correct skeleton if successfully deducing the right operator graph, thereby highly likely getting the accurate expressions as well. While OF-Net's $R^2$ performance was slightly inferior to NGGP, it surpassed both T-JSL and NeSymRes, and was comparable to GP. This was an unexpectedly positive result because methods like GP, progressively minimizing losses, can maintain high R2 values without identifying the correct expressions. In contrast, OF-Net, which is not driven by numerical loss, incurs more significant errors when operator graph are incorrect. Its commendable $R^2$ outcomes are, to a large extent, contingent upon the complete recovery leading to scores '1.0'.

\begin{figure}[h]
\centering
\includegraphics[width=0.9\columnwidth]{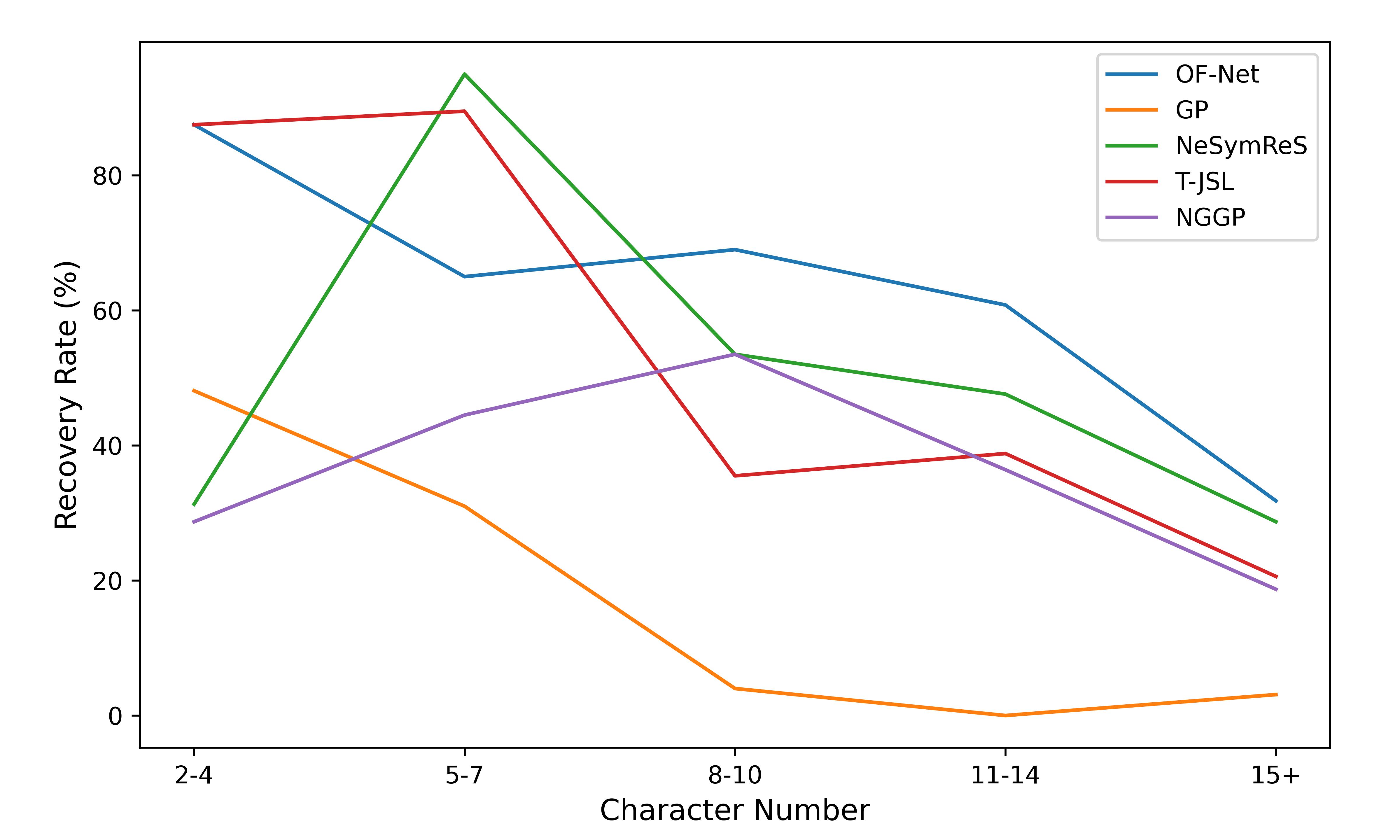}
\caption{Performance on different length}
\label{Result_length}
\end{figure}
According to the distribution of the test set, we evenly segmented the expression lengths, which is represented by character number, into five intervals and charted the performance of each method. Variance was calculated as
$$
Var = \frac{\sum_{i=1}^{5}n_i\times (\overline{x}_{i} - \overline{x})^2}{\sum_{i=1}^{5}n_i}
$$
where $n$ is expressions' number and $\overline{x}$ means the average. We get 0.023 for OF-Net second only to NGGP, which is satisfactory for a pre-trained model. The visual and variance indicate that OF-Net is little impacted by expression length, showcasing superior stability, suggesting that OF-Net possesses splendid extrapolative capabilities. For expressions exceeding 15 characters, a decline in performance was observed across all methods, largely due to the limitations in constant optimization. For instance, both OF-Net and NGGP encountered scenarios where the skeleton was accurately identified but the correct constants were not determined. Therefore, should ideal constant optimization be achievable, OF-Net's extrapolative potential could be further enhanced.

\subsection{Discussion}

OF-NET shows a significant difference in performance between univariate and bivariate data. It achieves exceptional results in univariate expressions, matching the optimal R2 of NGGP while leading significantly in recovery rates. Detailed analysis reveals that OF-NET almost always predicts the correct skeleton for univariate expressions, with only 8.8\% of errors due to constant optimization, mainly contributing to its outstanding performance. However, for bivariate data, OF-NET's predictions are less satisfactory, with lower R2 and recovery rates compared to other methods. Notably, 37.7\% of expressions lacked the correct skeleton, and 32\% of these errors were due to issues with operator prediction, which indicates the current limitations. We preliminarily attribute this issue to the training data. As bivariate operator structures are far more complex than univariate ones, it necessitate more training data. However, constrained by hardware limitations, we used the same data volume for both.

Addition and multiplication operators may introduce extraneous constants, which impedes constant optimization, resulting in certain non-minimal correct skeletons not being optimized successfully, and rendering some correct skeletons more complex. For instance, the standard solution for Nguyen-7 is $log(x_0+1)+log(pow(x_0,2)+1)$, but the expression (four decimal places) derived might be $log(5.2422\times x_0 + 5.2422) + log(0.1908\times pow(x_0,2) + 0.1908)$. Although mathematically equivalent, the redundant constants clearly complicate the expression, increasing the computational burden. Moreover, expressions like Koza-3 and Nguyen-4 encounter difficulties in optimizing the correct skeleton to an $R^2$ of 1.0 due to issues with complex numerical optimization of higher-order power.

Upon analysing results in greater detail, it becomes apparent that the range of variable exerts an influence on the outcomes. For instance, the structure $sin(x_0)\times cos(x_1)$ is prone to misjudgment when variables are in range (-1,1), leading to unsuccessful recovery for expressions involving this structure, such as Keijzer-13 and Nguyen-10. However, Livermore-10, which assumes a variable range of (0,1), is successfully identified.

According to the analysis of demerits, it is merited to take improvements in future work for superior performance. Specifically as follow:
\begin{itemize}
\item Enrich bivariable training dataset with broader diversified data.
\item Revise operator set to minimize overlap, thereby reducing the introduction of constants.
\item Propose an effective constant optimization algorithm especially in complex domain, tailored for the power operator.
\item Implement uniform scale transformation prior to executing the algorithm, followed by an inverse transformation on predicted expression.
\end{itemize}

\section{Conclusion}

In terms of the demerits of string predicting for the tree representation of expressions in symbolic regression, OF-Net uses the operator graph for representation and encodes the mathematical operations into feature codes, which are utilized instead of numerical value loss to discern the operator structure within unknown target expressions, thereby obtaining candidate skeletons. After constant optimizing, OF-Net can finally recover the expression. OF-Net utilizes the neural network fitting to represents the calculation process of the operators, in which multi-dimensional features of different operators are aggregated so as to extract more comprehensive and representative operator features.

The experiments are conducted on available public data sets and compared with four classical methods, GP, NeSymReS, T-JSL ans NGGP, involving every category of methods in symbolic regression. It confirms that OF-Net, compared to contrast methods, achieves a satisfactory performance with the highest recovery rate and excellent $R^2$, as well as less variance for expression lengths which leads to a splendid stability and extrapolative capabilities.

OF-Net is also possible for improvement in future work, anticipating refinement in dataset, operator set, constant optimization and pre-processing.

\begin{table*}[t]
\caption{Univariate Test Set}
\begin{tabular*}{\textwidth}{@{\extracolsep\fill}lcc}
\toprule
Benchmark & Expression & Range\\
\midrule
Keijzer-3 & $0.3 \times x_1 \times sin(2 \times pi \times x_1)$ & (-1,1)\\
Keijzer-4 & $pow(x_1,3) \times exp(-x_1) \times cos(x_1) \times sin(x_1)\times$ & (-1,1)\\
\, & $(pow(sin(x_1),2) \times cos(x_1)-1)$ & \,\\
Keijzer-6 & $0.5 \times x_1 \times (x_1+1)$ & (-1,1)\\
Keijzer-7 & $log(x_1)$ & (0.01,5)\\
Keijzer-8 & $sqrt(x_1)$ & (0,1)\\
Keijzer-9 & $log(x_1+sqrt(pow(x_1,2)+1))$ & (-1,1)\\
Koza-2 & $pow(x_1,5)-2 \times pow(x_1,3)+x_1$& (-1,1)\\
Koza-3 & $pow(x_1,6)-2 \times pow(x_1,4)+pow(x_1,2)$ & (-1,1\\
Nguyen-1 & $pow(x_1,3)+pow(x_1,2)+x_1$ & (-1,1)\\
Nguyen-2 & $pow(x_1,4)+pow(x_1,3)+pow(x_1,2)+x_1$ & (-1,1)\\
Nguyen-3 & $pow(x_1,5)+pow(x_1,4)+pow(x_1,3)+pow(x_1,2)+x_1$ & (-1,1)\\
Nguyen-4 & $pow(x_1,6)+pow(x_1,5)+pow(x_1,4)+pow(x_1,3)+pow(x_1,2)+x_1$ & (-1,1)\\
Nguyen-5 & $sin(pow(x_1,2)) \times cos(x_1)-1$ & (-1,1)\\
Nguyen-6 & $sin(x_1)+sin(x_1+pow(x_1,2))$ & (-1,1)\\
Nguyen-7 & $log(x_1+1)+log(pow(x_1,2)+1)$ & (0,2)\\
Nguyen-8 & $sqrt(x_1)$ & (0,4)\\
Constant-1 & $3.39 \times pow(x_1,3)+2.12 \times pow(x_1,2)+1.78 \times x_1$ & (-1,1)\\
Constant-2 & $sin(pow(x_1,2)) \times cos(x_1)-0.75$ & (-1,1)\\
Constant-5 & $sqrt(1.23 \times x_1)$ & (0,4)\\
Constant-6 & $pow(x_1,0.426)$ & (0,2)\\
Constant-8 & $log(x_1+1.4)+log(pow(x_1,2)+1.3)$ & (-1,1)\\
Livermore-1 & $1/3+x_1+sin(pow(x_1,2))$ & (-2,2)\\
Livermore-2 & $sin(pow(x_1,2)) \times cos(x_1)-2$ & (-1,1)\\
Livermore-3 & $sin(pow(x_1,3)) \times cos(pow(x_1,2))-1$ & (-1,1)\\
Livermore-4 & $log(x_1+1)+log(pow(x_1,2)+1)+log(x_1)$ & (0,1)\\
Livermore-6 & $4 \times pow(x_1,4)+3 \times pow(x_1,3)+2 \times pow(x_1,2)+x_1$ & (-1,1)\\
Livermore-7 & $0.5 \times (exp(x_1)-exp(-1 \times x_1))$ & (-1,1)\\
Livermore-8 & $0.5 \times (exp(x_1)+exp(-1 \times x_1))$ & (-1,1)\\
Livermore-9 & $pow(x_1,9)+pow(x_1,8)+pow(x_1,7)+pow(x_1,6)+pow(x_1,5)+$ & (-1,1)\\
\, & $pow(x_1,4)+pow(x_1,3)+pow(x_1,2)+x_1$ & \,\\
Livermore-13 & $pow(x_1,1/3)$ & (0,1)\\
Livermore-15 & $pow(x_1,1/5)$ & (0,1)\\
Livermore-16 & $pow(x_1,2/3)$ & (0,1)\\
Livermore-18 & $sin(pow(x_1,2)) \times cos(x_1)-5$ & (-1,1)\\
Livermore-19 & $pow(x_1,5)+pow(x_1,4)+pow(x_1,2)+x_1$ & (-1,1)\\
Livermore-20 & $exp(-1 \times pow(x_1,2))$ & (-1,1)\\
Livermore-21 & $pow(x_1,8)+pow(x_1,7)+pow(x_1,6)+pow(x_1,5)$+ & (-1,1)\\
\, & $pow(x_1,4)+pow(x_1,3)+pow(x_1,2)+x_1$ & \,\\
Livermore-22 & $exp(-0.5 \times pow(x_1,2))$ & (-1,1)\\
R1 & $\frac{pow(x_1+1,3)}{pow(x_1,2)-x_1+1}$ & (-1,1)\\
R2 & $\frac{(pow(x_1,5)-3 \times pow(x_1,3)+1)}{pow(x_1,2)+1)}($ & (-1,1)\\
R3 & $\frac{pow(x_1,6)+pow(x_1,5)}{pow(x_1,4)+pow(x_1,3)+pow(x_1,2)+x_1+1}$ & (-1,1)\\
Nguyen-1c & $3.39 \times pow(x_1,3)+2.12 \times pow(x_1,2)+1.78 \times x_1$ & (-1,1)\\
Nguyen-5c & $sin(pow(x_1,2 \times cos(x_1)-0.75$ & (-1,1)\\
Nguyen-7c & $log(x_1+1.4)+log(pow(x_1,2)+1.3)$ & (-1,1)\\
Nguyen-8c & $sqrt(1.23 \times x_1)$ & (0,1)\\
Feynman I.6.2a & $\frac{0.5 \times exp(-pow(x_1,2))}{\sqrt{2 \times pi}} $ & (1,5)\\
\hline
\end{tabular*}
\label{Univariate Test Set}
\end{table*}


\begin{table*}[t]
\caption{Bivariate Test Set}
\begin{tabular*}{\textwidth}{@{\extracolsep\fill}lcc}
\toprule
Benchmark & Expression & Range\\
\midrule
Keijzer-10 & $pow(x_1,x_2)$ & (0,1);(0,1)\\
Keijzer-11 & $x_1 \times x_2+sin((x_1-1) \times (x_2-1))$ & (-1,1);(-1,1)\\
Keijzer-12 & $pow(x_1,4)-pow(x_1,3)+div(pow(x_2,2),2)-x_2$ & (-1,1);(-1,1)\\
Keijzer-13 & $6 \times sin(x_1) \times cos(x_2)$ & (-1,1);(-1,1)\\
Keijzer-14 & $\frac{8}{2+pow(x_1,2)+pow(x_2,2)}$ & (-1,1);(-1,1)\\
Keijzer-15 & $0.2 \times pow(x_1,3)+0.5 \times pow(x_2,3)-x_2-x_1$ & (-1,1);(-1,1)\\
Nguyen-9 & $sin(x_1)+sin(pow(x_2,2))$ & (-1,1);(-1,1)\\
Nguyen-10 & $2 \times sin(x_1) \times cos(x_2)$ & (-1,1);(-1,1)\\
Nguyen-11 & $pow(x_1,x_2)$ & (0,1);(0,1)\\
Nguyen-12 & $pow(x_1,4)-pow(x_1,3)+0.5 \times pow(x_2,2)-x_2$ & (-1,1);(-1,1)\\
Constant-3 & $sin(1.5 \times x_1) \times cos(0.5 \times x_2)$ & (-1,1);(-1,1)\\
Constant-4 & $2.7 \times pow(x_1,x_2)$ & (0,1);(0,1)\\
Constant-7 & $2 \times sin(1.3 \times x_1) \times cos(x_2)$ & (0,1);(0,1)\\
Livermore-5 & $pow(x_1,4)-pow(x_1,3)+pow(x_2,2)-x_2$ & (0,1);(0,1)\\
Livermore-10 & $6 \times sin(x_1) \times cos(x_2)$ & (0,1);(0,1)\\
Livermore-11 & $\frac{pow(x_1,2) \times pow(x_2,2)}{x_1+x_2}$ & (-2,-0.1 and 0.1,2)(-2,-0.1 and 0.1,2);\\
Livermore-12 & $\frac{pow(x1,5)}{pow(x2,3)}$ & (-1,1);(-2,-0.5 and 0.5,2)\\
Livermore-14 & $pow(x_1,3)+pow(x_1,2)+x_1+$ & (-1,1);(-1,1)\\
\, & $sin(x_1)+sin(pow(x_2,2))$ & \,\\
Livermore-17 & $4 \times sin(x_1) \times cos(x_2)$ & (-1,1);(-1,1)\\
Nguyen-10c & $sin(1.5 \times x_1) \times cos(0.5 \times x_2)$ & (-1,1);(-1,1)\\
Jin-1 & $2.5 \times pow(x_1,4) - 1.3 \times pow(x_1,3)+$ & (-1,1);(-1,1)\\
\, & $0.5 \times pow(x_2,2) - 1.7 \times x_2$ & \,\\
Jin-2 & $8.0 \times pow(x_1,2) + 8.0 \times pow(x_2,3) -15.0$ & (-1,1);(-1,1)\\
Jin-3 & $0.2 \times pow(x_1,3) + 0.5 \times pow(x_2,3) -$ & (-1,1);(-1,1)\\
\, & $ 1.2 \times x_2 - 0.5 \times x_1$ & \,\\
Jin-4 & $1.5 \times exp(x_1) + 5.0 \times cos(x_2)$ & (-1,1);(-1,1)\\
Jin-5 & $6.0 \times sin(x_1) \times cos(x_2)$ & (-1,1);(-1,1)\\
Jin-6 & $1.35 \times x_1 \times x_2 + 5.5 \times sin((x_1-1.0) \times (x_2-1.0))$ & (-1,1);(-1,1)\\
bacterial respiration-1 & $20-x_1-\frac{x_1 \times x_2}{1+0.5 \times pow(x_1,2)}$ & (-1,1);(-1,1)\\
bacterial respiration-3 & $10-\frac{x_1 \times x_2}{1+0.5 \times pow(x_1,2)}$ & (-1,1);(-1,1)\\
bar magnets-1 & $0.5 \times sin(x_1-x_2)-sin(x_1)$ & (-1,1);(-1,1)\\
bar magnets-2 & $0.5 \times sin(x_2-x_1)-sin(x_2)$ & (-1,1);(-1,1)\\
glider-1 & $-0.05 \times pow(x_1,2)-sin(x_2)$ & (-1,1);(-1,1)\\
glider-2 & $x_1-\frac{cos(x_2)}{x_1}$ & (-2,-0.1 and 0.1,2);(-1,1)\\
Lotka-Volterra-1 & $3 \times x_1-2 \times x_1 \times x \times 2-pow(x_1,2)$ & (-1,1);(-1,1)\\
Lotka-Volterra-2 & $2 \times x_2-x_1 \times x_2-pow(x_2,2)$ & (-1,1);(-1,1)\\
predator-prey-1 & $x_1 \times (4-x_1-\frac{x_2}{1+x_1})$ & (-0.9,0.9);(-1,1)\\
predator-prey-2 & $x_2 \times (\frac{x_1}{1+x_1}-0.075 \times x_2)$ & (-0.9,0.9);(-1,1)\\
shear flow-1 & $cot(x_2) \times cos(x_1)$ & (-2,-0.1 and 0.1,2)(-2,-0.1 and 0.1,2);\\
shear flow-2 & $(pow(cos(x_2),2)+$  & (-1,1);(-1,1)\\
\, & $0.1 \times pow(sin(x_2),2)) \times sin(x_1)$ & \,\\
van der Pol-1 & $10 \times (x_2-(1/3 \times (pow(x_1,3)-x_1)))$ & (-1,1);(-1,1)\\
van der Pol-2 & $-0.1 \times x_1$ & (-1,1);(-1,1)\\
Feynman I.6.2 & $\frac{exp(-0.5 \times pow((x_2/x_1),2))}{\sqrt{2 \times pi \times x1}}$ & (1,5);(1,5)\\
Feynman I.12.1 & $x_1 \times x_2$ & (1,5);(1,5)\\
Feynman I.12.5 & $x_1 \times x_2$ & (1,5);(1,5)\\
Feynman I.14.4 & $0.5 \times x_1 \times pow(x_2,2)$ & (-3,3);(-3,3)\\
Feynman I.25.13 & $\frac{x_1}{x_2}$ & (1,5);(1,5)\\
Feynman I.29.4 & $\frac{x_1}{x_2}$ & (1,5);(1,5)\\
Feynman I.34.27 & $\frac{x_2/}{2 \times pi} \times x_1$ & (1,5);(1,5)\\
Feynman I.39.1 & $1.5 \times x_1 \times x_2$ & (-3,3);(-3,3)\\
Feynman II.8.31 & $0.5 \times x_1 \times pow(x_2,2)$ & (-3,3);(-3,3)\\
Feynman II.11.28 & $1+x_1 \times \frac{x_2}{1-\frac{x_1 \times x_2}{3}}$ & (-5,-2 and 2,5)(-5,-2 and 2,5);\\
Feynman II.27.18 & $x_1 \times pow(x_2,2)$ & (-2,2);(-2,2)\\
Feynman II.38.14 & $\frac{x_1}{2 \times (1+x_2)}$ & (1,5);(1,5)\\
Feynman III.12.43 & $x_1 \times \frac{x_2}{2 \times pi}$ & (1,5);(1,5)\\
\hline
\end{tabular*}
\label{Bivariate Test Set}
\end{table*}

\begin{table*}[t]
\caption{Detailed Result of Univariate Test Set}
\begin{tabular*}{\textheight}{lcccccccccc}
\toprule%
& \multicolumn{2}{c}{OF-Net} & \multicolumn{2}{c}{GP} & \multicolumn{2}{c}{NeSymReS} & \multicolumn{2}{c}{T-JSL} & \multicolumn{2}{c}{DSO}\\\cmidrule{2-3}\cmidrule{4-5}\cmidrule{6-7}\cmidrule{8-9}\cmidrule{10-11}%
Benchmark & $R^2$	& Recovery & $R^2$	& Recovery  & $R^2$	& Recovery  & $R^2$	& Recovery  & $R^2$	& Recovery \\
\midrule
Keijzer-3 & 1 & 1 & 0.691 & 0 & 0.386 & 0 & 1 & 0.2 & 0.977 & 0\\
Keijzer-4 & 1 & 0.1 & 0.526 & 0 & 0.942 & 0 & 0.398 & 0 & 0.999 & 0\\
Keijzer-6 & 1 & 1 & 0.977 & 0 & 1 & 1 & 1 & 1 & 1 & 1\\
Keijzer-7 & 1 & 1 & 1 & 1 & None & 0 & 1 & 1 & 1 & 1\\
Keijzer-8 & 1 & 1 & 1 & 1 & None & 0 & 1 & 1 & 0.999 & 0\\
Keijzer-9 & 1 & 1 & 0.999 & 0 & 0.999 & 0 & 0.999 & 0 & 1 & 1\\
Koza-2 & 1 & 1 & 0.992 & 0 & 1 & 1 & 0.945 & 0 & 1 & 1\\
Koza-3 & 0.906 & 0 & 0.259 & 0 & 0.989 & 0 & 0.878 & 0 & 1 & 0.9\\
Nguyen-1 & 1 & 1 & 1 & 0.2 & 1 & 1 & 1 & 1 & 1 & 1\\
Nguyen-2 & 1 & 1 & 0.993 & 0 & 1 & 1 & 1 & 1 & 1 & 1\\
Nguyen-3 & 1 & 0.9 & 0.997 & 0 & 1 & 0.9 & 0.836 & 0 & 0.999 & 0\\
Nguyen-4 & 0.952 & 0 & 0.992 & 0 & 0.256 & 0 & -1.512 & 0 & 0.999 & 0\\
Nguyen-5 & 1 & 1 & 0.726 & 0 & 1 & 1 & 0.048 & 0 & 1 & 1\\
Nguyen-6 & 0.999 & 0 & 0.998 & 0 & 1 & 1 & 1 & 0.9 & 1 & 1\\
Nguyen-7 & 1 & 1 & 0.878 & 0 & None & 0 & None & 0 & 1 & 1\\
Nguyen-8 & 1 & 1 & 0.967 & 0 & None & 0 & 1 & 1 & 1 & 0.6\\
Constant-1 & 1 & 1 & 1 & 0.5 & 1 & 0.9 & 1 & 1 & 1 & 1\\
Constant-2 & 1 & 1 & 0.973 & 0 & 0.012 & 0 & 0.083 & 0 & 1 & 0.9\\
Constant-5 & 1 & 1 & 0.999 & 0 & 0.116 & 0 & 1 & 1 & 0.999 & 0\\
Constant-6 & 1 & 1 & 0.982 & 0 & None & 0 & 1 & 1 & 0.997 & 0\\
Constant-8 & 1 & 1 & 0.991 & 0 & None & 0 & 0.062 & 0 & 1 & 0.9\\
Livermore-1 & 1 & 0.8 & 0.989 & 0 & 1 & 1 & 1 & 1 & 1 & 1\\
Livermore-2 & 1 & 1 & 0.971 & 0 & 1 & 1 & 0.585 & 0 & 1 & 1\\
Livermore-3 & 1 & 1 & 0.991 & 0 & None & 0 & -0.707 & 0 & 1 & 0.8\\
Livermore-4 & 1 & 0.5 & 0.999 & 0 & None & 0 & 0.999 & 0 & 0.995 & 0\\
Livermore-6 & 1 & 1 & 0.984 & 0 & 1 & 1 & 1 & 1 & 1 & 1\\
Livermore-7 & 0.999 & 0 & 0.999 & 0 & 0.999 & 0 & 0.999 & 0 & 1 & 1\\
Livermore-8 & 0.924 & 0 & 0.999 & 0 & 1 & 0.5 & 1 & 1 & 1 & 1\\
Livermore-9 & 0.997 & 0 & 0.981 & 0 & 0.289 & 0 & -7.754 & 0 & 0.999 & 0\\
Livermore-13 & 1 & 1 & 0.993 & 0 & None & 0 & 1 & 1 & 0.999 & 0\\
Livermore-15 & 1 & 1 & 0.938 & 0 & None & 0 & 1 & 1 & 0.998 & 0\\
Livermore-16 & 1 & 1 & 0.987 & 0 & None & 0 & 1 & 1 & 0.999 & 0\\
Livermore-18 & 1 & 1 & 0.581 & 0 & 0.999 & 0 & 0.025 & 0 & 0.999 & 0\\
Livermore-19 & 1 & 1 & 0.975 & 0 & 1 & 1 & 0.973 & 0 & 1 & 0.4\\
Livermore-20 & 1 & 1 & 0.972 & 0 & 1 & 1 & 1 & 1 & 1 & 1\\
Livermore-21 & 0.998 & 0 & 0.995 & 0 & 1 & 0.3 & -8.625 & 0 & 0.999 & 0\\
Livermore-22 & 1 & 1 & 0.989 & 0 & 1 & 1 & 1 & 1 & 1 & 1\\
R1 & 1 & 1 & 0.994 & 0 & 1 & 0.8 & 1 & 1 & 1 & 1\\
R2 & 1 & 1 & 0.916 & 0 & 0.999 & 0 & 0.997 & 0 & 0.999 & 0\\
R3 & 0.999 & 0 & 0.818 & 0 & 0.999 & 0 & 0.999 & 0 & 0.998 & 0\\
Nguyen-1c & 1 & 1 & 0.995 & 0 & 1 & 1 & 1 & 1 & 1 & 1\\
Nguyen-5c & 1 & 1 & 0.892 & 0 & 0.009 & 0 & 0.063 & 0 & 1 & 1\\
Nguyen-7c & 1 & 1 & 0.986 & 0 & \ & 0 & 1 & 0.4 & 1 & 1\\
Nguyen-8c & 1 & 1 & 0.991 & 0 & \ & 0 & 1 & 1 & 0.999 & 0\\
Feynman I.6.2a & 0.999 & 0 & 0.886 & 0 & 1 & 1 & 1 & 1 & 1 & 1\\
\hline
\end{tabular*}
\end{table*}

\begin{table*}[t]
\caption{Detailed Result of Bivariate Test Set}
\begin{tabular*}{\textheight}{lcccccccccc}
\toprule%
& \multicolumn{2}{c}{OF-Net} & \multicolumn{2}{c}{GP} & \multicolumn{2}{c}{NeSymReS} & \multicolumn{2}{c}{T-JSL} & \multicolumn{2}{c}{DSO}\\\cmidrule{2-3}\cmidrule{4-5}\cmidrule{6-7}\cmidrule{8-9}\cmidrule{10-11}%
Benchmark & $R^2$	& Recovery & $R^2$	& Recovery  & $R^2$	& Recovery  & $R^2$	& Recovery  & $R^2$	& Recovery \\
\midrule
Keijzer-10 & 1 & 1 & 1 & 0.5 & None & 0 & 0.097 & 0 & 0.988 & 0\\
Keijzer-11 & 1 & 1 & 0.274 & 0 & 0.999 & 0 & 0.999 & 0 & 0.948 & 0\\
Keijzer-12 & 0.051 & 0 & 0.61 & 0 & 0.999 & 0 & 0.999 & 0 & 0.949 & 0\\
Keijzer-13 & 0.958 & 0 & 0.963 & 0 & 1 & 1 & 1 & 1 & 0.999 & 0\\
Keijzer-14 & 1 & 1 & 0.896 & 0 & 1 & 0.6 & 0.987 & 0 & 0.971 & 0\\
Keijzer-15 & 0.982 & 0 & 0.817 & 0 & 1 & 1 & 0.999 & 0 & 0.994 & 0\\
Nguyen-9 & None & 0 & 1 & 0.4 & 1 & 1 & 1 & 1 & 0.998 & 0\\
Nguyen-10 & None & 0 & 1 & 0.4 & 1 & 1 & 1 & 1 & 0.999 & 0\\
Nguyen-11 & 1 & 1 & 1 & 0.4 & None & 0 & -0.149 & 0 & 0.988 & 0\\
Nguyen-12 & 0.045 & 0 & 0.541 & 0 & 1 & 0.4 & 1 & 0.3 & 0.966 & 0\\
Constant-3 & 1 & 1 & 0.998 & 0 & 0.013 & 0 & 0 & 0 & 1 & 0.8\\
Constant-4 & 1 & 1 & 0.655 & 0 & None & 0 & -0.057 & 0 & 0.981 & 0\\
Constant-7 & 1 & 1 & 0.989 & 0 & 1 & 1 & -0.226 & 0 & 0.999 & 0\\
Livermore-5 & None & 0 & 0.822 & 0 & 1 & 1 & 1 & 1 & 0.984 & 0\\
Livermore-10 & 1 & 1 & 0.985 & 0 & 1 & 1 & 1 & 1 & 1 & 0.9\\
Livermore-11 & 1 & 1 & 0.429 & 0 & 1 & 0.7 & 1 & 1 & 0.995 & 0\\
Livermore-12 & 1 & 1 & 0.983 & 0 & 1 & 1 & 0.088 & 0 & 0.991 & 0\\
Livermore-14 & 0.734 & 0 & 0.994 & 0 & 0.999 & 0 & 0.999 & 0 & 0.999 & 0\\
Livermore-17 & 0.968 & 0 & 0.974 & 0 & 1 & 1 & 1 & 1 & 0.999 & 0\\
Nguyen-10c & 1 & 1 & 0.998 & 0 & 0.039 & 0 & 0.006 & 0 & 0.999 & 0\\
Jin-1 & 0.279 & 0 & 0.764 & 0 & 0.999 & 0 & 0.999 & 0 & 0.934 & 0\\
Jin-2 & 1 & 1 & 0.845 & 0 & 1 & 0.7 & 1 & 1 & 0.905 & 0\\
Jin-3 & 0.976 & 0 & 0.975 & 0 & 0.999 & 0 & 0.999 & 0 & 0.991 & 0\\
Jin-4 & 1 & 1 & 0.972 & 0 & 1 & 1 & 1 & 1 & 0.963 & 0\\
Jin-5 & 0.968 & 0 & 0.986 & 0 & 1 & 1 & 1 & 1 & 0.999 & 0\\
Jin-6 & 1 & 1 & 0.867 & 0 & 0.992 & 0 & 0.992 & 0 & 0.969 & 0\\
bacterial respiration 1 & 0.997 & 0 & 0.752 & 0 & 0.915 & 0 & 0.839 & 0 & 0.997 & 0\\
bacterial respiration 3 & None & 0 & -0.003 & 0 & 0.576 & 0 & 0.432 & 0 & 0.996 & 0\\
bar magnets 1 & 0.98 & 0 & 0.703 & 0 & 1 & 0.4 & 0.783 & 0 & 0.989 & 0\\
bar magnets 2 & 0.979 & 0 & 0.762 & 0 & 1 & 0.8 & 0.812 & 0 & 0.998 & 0\\
glider 1 & 0.998 & 0 & 0.998 & 0 & 1 & 1 & 1 & 1 & 0.999 & 0\\
glider 2 & 0.06	0 & 1 & 0.4 & 1 & 1 & 1 & 0.9 & 0.154 & 0\\
Lotka-Volterra 1 & 1 & 1 & 0.821 & 0 & 1 & 1 & 1 & 1 & 1 & 1\\
Lotka-Volterra 2 & 0.65 & 0 & 0.844 & 0 & 1 & 1 & 1 & 1 & 1 & 1\\
predator-prey 1 & 0.973 & 0 & 0.431 & 0 & 1 & 0.9 & 1 & 1 & 0.979 & 0\\
predator-prey 2 & 0.465 & 0 & 0.992 & 0 & 1 & 0.3 & 1 & 0.5 & 0.971 & 0\\
shear flow 1 & 1 & 1 & 1 & 0.6 & 1 & 1 & 1 & 1 & 0.949 & 0\\
shear flow 2 & 0.993 & 0 & 0.999 & 0 & 0.999 & 0 & -0.006 & 0 & 0.999 & 0\\
van der Pol 1 & 1 & 1 & 0.989 & 0 & 1 & 1 & 1 & 0.8 & 0.994 & 0\\
van der Pol 2 & 1 & 1 & 1 & 0.8 & 1 & 1 & 1 & 1 & 1 & 1\\
Feynman I.6.2 & 0.988 & 0 & 0.746 & 0 & 0.822 & 0 & 0.931 & 0 & 0.863 & 0\\
Feynman I.12.1 & 1 & 1 & 1 & 1 & 1 & 1 & 1 & 1 & 1 & 1\\
Feynman I.12.5 & 1 & 1 & 1 & 1 & 1 & 1 & 1 & 1 & 1 & 1\\
Feynman I.14.4 & 1 & 1 & 0.998 & 0 & 1 & 1 & 1 & 1 & 0.992 & 0\\
Feynman I.25.13 & 0.346 & 0 & 1 & 1 & 1 & 1 & 1 & 1 & 0.998 & 0\\
Feynman I.29.4 & 0.345 & 0 & 1 & 1 & 1 & 1 & 1 & 1 & 0.999 & 0\\
Feynman I.34.27 & 1 & 1 & 1 & 1 & 1 & 1 & 1 & 1 & 1 & 1\\
Feynman I.39.1 & 1 & 1 & 1 & 1 & 1 & 1 & 1 & 1 & 1 & 1\\
Feynman II.8.31 & 1 & 1 & 1 & 0.8 & 1 & 1 & 1 & 1 & 0.992 & 0\\
Feynman II.11.28 & 1 & 0.8 & 0.828 & 0 & 0.255 & 0 & 0.76 & 0 & 0.966 & 0\\
Feynman II.27.18 & 1 & 1 & 1 & 0.7 & 1 & 1 & 1 & 1 & 1 & 1\\
Feynman II.38.14 & 1 & 1 & 1 & 0.5 & 1 & 1 & 1 & 1 & 0.999 & 0\\
Feynman III.12.43 & 1 & 1 & 1 & 1 & 1 & 1 & 1 & 1 & 1 & 1\\
\hline
\end{tabular*}
\end{table*}

\bibliography{aaai25}

\begin{thebibliography}{26}
\providecommand{\natexlab}[1]{#1}

\bibitem[{Biggio et~al.(2021)Biggio, Bendinelli, Neitz, Lucchi, and Parascandolo}]{biggio2021neural}
Biggio, L.; Bendinelli, T.; Neitz, A.; Lucchi, A.; and Parascandolo, G. 2021.
\newblock Neural symbolic regression that scales.
\newblock In \emph{International Conference on Machine Learning}, 936--945. Pmlr.

\bibitem[{Chen et~al.(2023)Chen, Wang, Miao, Mo, Feng, Zhou, and Wang}]{chen2023transformer}
Chen, L.; Wang, Y.; Miao, Z.; Mo, Y.; Feng, M.; Zhou, Z.; and Wang, H. 2023.
\newblock Transformer-based imitative reinforcement learning for multi-robot path planning.
\newblock \emph{IEEE Transactions on Industrial Informatics}.

\bibitem[{Chen and Chen(1995)}]{chen1995universal}
Chen, T.; and Chen, H. 1995.
\newblock Universal approximation to nonlinear operators by neural networks with arbitrary activation functions and its application to dynamical systems.
\newblock \emph{IEEE transactions on neural networks}, 6(4): 911--917.

\bibitem[{Devlin et~al.(2018)Devlin, Chang, Lee, and Toutanova}]{devlin2018bert}
Devlin, J.; Chang, M.-W.; Lee, K.; and Toutanova, K. 2018.
\newblock Bert: Pre-training of deep bidirectional transformers for language understanding.
\newblock \emph{arXiv preprint arXiv:1810.04805}.

\bibitem[{Holt, Qian, and van~der Schaar(2023)}]{holt2023deep}
Holt, S.; Qian, Z.; and van~der Schaar, M. 2023.
\newblock Deep generative symbolic regression.
\newblock \emph{arXiv preprint arXiv:2401.00282}.

\bibitem[{Jin, Meng, and Lu(2022)}]{jin2022mionet}
Jin, P.; Meng, S.; and Lu, L. 2022.
\newblock MIONet: Learning multiple-input operators via tensor product.
\newblock \emph{SIAM Journal on Scientific Computing}, 44(6): A3490--A3514.

\bibitem[{Kamienny et~al.(2022)Kamienny, d'Ascoli, Lample, and Charton}]{kamienny2022end}
Kamienny, P.-A.; d'Ascoli, S.; Lample, G.; and Charton, F. 2022.
\newblock End-to-end symbolic regression with transformers.
\newblock \emph{Advances in Neural Information Processing Systems}, 35: 10269--10281.

\bibitem[{Landajuela et~al.(2022)Landajuela, Lee, Yang, Glatt, Santiago, Aravena, Mundhenk, Mulcahy, and Petersen}]{landajuela2022unified}
Landajuela, M.; Lee, C.~S.; Yang, J.; Glatt, R.; Santiago, C.~P.; Aravena, I.; Mundhenk, T.; Mulcahy, G.; and Petersen, B.~K. 2022.
\newblock A unified framework for deep symbolic regression.
\newblock \emph{Advances in Neural Information Processing Systems}, 35: 33985--33998.

\bibitem[{Lee et~al.(2019)Lee, Lee, Kim, Kosiorek, Choi, and Teh}]{lee2019set}
Lee, J.; Lee, Y.; Kim, J.; Kosiorek, A.; Choi, S.; and Teh, Y.~W. 2019.
\newblock Set transformer: A framework for attention-based permutation-invariant neural networks.
\newblock In \emph{International conference on machine learning}, 3744--3753. PMLR.

\bibitem[{Li et~al.(2022)Li, Li, Sun, Wu, Yu, Liu, Li, and Tian}]{li2022transformer}
Li, W.; Li, W.; Sun, L.; Wu, M.; Yu, L.; Liu, J.; Li, Y.; and Tian, S. 2022.
\newblock Transformer-based model for symbolic regression via joint supervised learning.
\newblock In \emph{The Eleventh International Conference on Learning Representations}.

\bibitem[{Li et~al.(2024)Li, Li, Yu, Wu, Liu, Li, Hao, Wei, and Deng}]{li2024discovering}
Li, Y.; Li, W.; Yu, L.; Wu, M.; Liu, J.; Li, W.; Hao, M.; Wei, S.; and Deng, Y. 2024.
\newblock Discovering Mathematical Formulas from Data via GPT-guided Monte Carlo Tree Search.
\newblock \emph{arXiv preprint arXiv:2401.14424}.

\bibitem[{Liu et~al.(2023)Liu, Li, Yu, Wu, Sun, Li, and Li}]{liu2023snr}
Liu, J.; Li, W.; Yu, L.; Wu, M.; Sun, L.; Li, W.; and Li, Y. 2023.
\newblock SNR: Symbolic network-based rectifiable learning framework for symbolic regression.
\newblock \emph{Neural Networks}, 165: 1021--1034.

\bibitem[{Lu, Jin, and Karniadakis(2019)}]{lu2019deeponet}
Lu, L.; Jin, P.; and Karniadakis, G.~E. 2019.
\newblock Deeponet: Learning nonlinear operators for identifying differential equations based on the universal approximation theorem of operators.
\newblock \emph{arXiv preprint arXiv:1910.03193}.

\bibitem[{Lu et~al.(2022{\natexlab{a}})Lu, Meng, Cai, Mao, Goswami, Zhang, and Karniadakis}]{lu2022comprehensive}
Lu, L.; Meng, X.; Cai, S.; Mao, Z.; Goswami, S.; Zhang, Z.; and Karniadakis, G.~E. 2022{\natexlab{a}}.
\newblock A comprehensive and fair comparison of two neural operators (with practical extensions) based on fair data.
\newblock \emph{Computer Methods in Applied Mechanics and Engineering}, 393: 114778.

\bibitem[{Lu et~al.(2022{\natexlab{b}})Lu, Pestourie, Johnson, and Romano}]{lu2022multifidelity}
Lu, L.; Pestourie, R.; Johnson, S.~G.; and Romano, G. 2022{\natexlab{b}}.
\newblock Multifidelity deep neural operators for efficient learning of partial differential equations with application to fast inverse design of nanoscale heat transport.
\newblock \emph{Physical Review Research}, 4(2): 023210.

\bibitem[{Mao et~al.(2021)Mao, Lu, Marxen, Zaki, and Karniadakis}]{mao2021deepm}
Mao, Z.; Lu, L.; Marxen, O.; Zaki, T.~A.; and Karniadakis, G.~E. 2021.
\newblock DeepM\&Mnet for hypersonics: Predicting the coupled flow and finite-rate chemistry behind a normal shock using neural-network approximation of operators.
\newblock \emph{Journal of computational physics}, 447: 110698.

\bibitem[{Meidani et~al.(2023)Meidani, Shojaee, Reddy, and Farimani}]{meidani2023snip}
Meidani, K.; Shojaee, P.; Reddy, C.~K.; and Farimani, A.~B. 2023.
\newblock SNIP: Bridging Mathematical Symbolic and Numeric Realms with Unified Pre-training.
\newblock \emph{arXiv preprint arXiv:2310.02227}.

\bibitem[{Mundhenk et~al.(2021)Mundhenk, Landajuela, Glatt, Santiago, Faissol, and Petersen}]{mundhenk2021symbolic}
Mundhenk, T.~N.; Landajuela, M.; Glatt, R.; Santiago, C.~P.; Faissol, D.~M.; and Petersen, B.~K. 2021.
\newblock Symbolic regression via neural-guided genetic programming population seeding.
\newblock \emph{arXiv preprint arXiv:2111.00053}.

\bibitem[{Petersen et~al.(2019)Petersen, Landajuela, Mundhenk, Santiago, Kim, and Kim}]{petersen2019deep}
Petersen, B.~K.; Landajuela, M.; Mundhenk, T.~N.; Santiago, C.~P.; Kim, S.~K.; and Kim, J.~T. 2019.
\newblock Deep symbolic regression: Recovering mathematical expressions from data via risk-seeking policy gradients.
\newblock \emph{arXiv preprint arXiv:1912.04871}.

\bibitem[{Sun et~al.(2022)Sun, Liu, Wang, and Sun}]{sun2022symbolic}
Sun, F.; Liu, Y.; Wang, J.-X.; and Sun, H. 2022.
\newblock Symbolic physics learner: Discovering governing equations via monte carlo tree search.
\newblock \emph{arXiv preprint arXiv:2205.13134}.

\bibitem[{Udrescu et~al.(2020)Udrescu, Tan, Feng, Neto, Wu, and Tegmark}]{udrescu2020ai2}
Udrescu, S.-M.; Tan, A.; Feng, J.; Neto, O.; Wu, T.; and Tegmark, M. 2020.
\newblock AI Feynman 2.0: Pareto-optimal symbolic regression exploiting graph modularity.
\newblock \emph{Advances in Neural Information Processing Systems}, 33: 4860--4871.

\bibitem[{Udrescu and Tegmark(2020)}]{udrescu2020ai1}
Udrescu, S.-M.; and Tegmark, M. 2020.
\newblock AI Feynman: A physics-inspired method for symbolic regression.
\newblock \emph{Science Advances}, 6(16): eaay2631.

\bibitem[{Vastl et~al.(2024)Vastl, Kulh{\'a}nek, Kubal{\'\i}k, Derner, and Babu{\v{s}}ka}]{vastl2024symformer}
Vastl, M.; Kulh{\'a}nek, J.; Kubal{\'\i}k, J.; Derner, E.; and Babu{\v{s}}ka, R. 2024.
\newblock Symformer: End-to-end symbolic regression using transformer-based architecture.
\newblock \emph{IEEE Access}.

\bibitem[{Vaswani et~al.(2017)Vaswani, Shazeer, Parmar, Uszkoreit, Jones, Gomez, Kaiser, and Polosukhin}]{vaswani2017attention}
Vaswani, A.; Shazeer, N.; Parmar, N.; Uszkoreit, J.; Jones, L.; Gomez, A.~N.; Kaiser, {\L}.; and Polosukhin, I. 2017.
\newblock Attention is all you need.
\newblock \emph{Advances in neural information processing systems}, 30.

\bibitem[{Wu et~al.(2023)Wu, Li, Yu, Sun, Liu, and Li}]{wu2023discovering}
Wu, M.; Li, W.; Yu, L.; Sun, L.; Liu, J.; and Li, W. 2023.
\newblock Discovering Mathematical Expressions Through DeepSymNet: A Classification-Based Symbolic Regression Framework.
\newblock \emph{IEEE Transactions on Neural Networks and Learning Systems}.

\bibitem[{Xu, Liu, and Sun(2023)}]{xu2023rsrm}
Xu, Y.; Liu, Y.; and Sun, H. 2023.
\newblock RSRM: Reinforcement Symbolic Regression Machine.
\newblock \emph{arXiv preprint arXiv:2305.14656}.

\end{thebibliography}

\end{document}